\def\year{2021}\relax
\documentclass[letterpaper]{article} % DO NOT CHANGE THIS
\usepackage{aaai21}  % DO NOT CHANGE THIS
\usepackage{times}  % DO NOT CHANGE THIS
\usepackage{helvet} % DO NOT CHANGE THIS
\usepackage{courier}  % DO NOT CHANGE THIS
\usepackage[hyphens]{url}  % DO NOT CHANGE THIS
\usepackage{graphicx} % DO NOT CHANGE THIS
\usepackage{graphicx}  % DO NOT CHANGE THIS
\usepackage{natbib}  % DO NOT CHANGE THIS AND DO NOT ADD ANY OPTIONS TO IT
\usepackage{caption} % DO NOT CHANGE THIS AND DO NOT ADD ANY OPTIONS TO IT
\usepackage{amsmath}
\usepackage{amsthm}
\usepackage{multirow}
\usepackage{booktabs}
\usepackage{algorithm}
\usepackage{algorithmic}
\usepackage{bm}
\usepackage{amssymb}% http://ctan.org/pkg/amssymb
\usepackage{pifont}% http://ctan.org/pkg/pifont
\newcommand{\cmark}{\ding{52}}%
\newcommand{\xmark}{\ding{56}}%
\title{Spatiotemporal Graph Neural Network based Mask Reconstruction for \\ Video Object Segmentation}

\author{Daizong Liu\textsuperscript{\rm 1}, Shuangjie Xu\textsuperscript{\rm 2}, Xiao-Yang Liu\textsuperscript{\rm 3}, Zichuan Xu\textsuperscript{\rm 4}, Wei Wei\textsuperscript{\rm 1}, Pan Zhou\textsuperscript{\rm 1*}\\
\textsuperscript{\rm 1}{\rm \normalsize Huazhong University of Science and Technology} \
\textsuperscript{\rm 2}{\rm \normalsize DEEPROUTE.AI}\\
\textsuperscript{\rm 3}{\rm \normalsize Columbia University} \
\textsuperscript{\rm 4}{\rm \normalsize Dalian University of Technology} \\
{\rm \normalsize \{dzliu, weiw, panzhou\}@hust.edu.cn, shuangjiexu@deeproute.ai, xl2427@columbia.edu, z.xu@dlut.edu.cn}
}

\begin{document}
\maketitle

\begin{abstract}
This paper addresses the task of segmenting class-agnostic objects in semi-supervised setting. Although previous detection based methods achieve relatively good performance, these approaches extract the best proposal by a greedy strategy, which may lose the local patch details outside the chosen candidate. In this paper, we propose a novel spatiotemporal graph neural network (STG-Net) to reconstruct more accurate masks for video object segmentation, which captures the local contexts by utilizing all proposals. In the spatial graph, we treat object proposals of a frame as nodes and represent their correlations with an edge weight strategy for mask context aggregation. To capture temporal information from previous frames, we use a memory network to refine the mask of current frame by retrieving historic masks in a temporal graph. The joint use of both local patch details and temporal relationships allow us to better address the challenges such as object occlusion and missing. Without online learning and fine-tuning, our STG-Net achieves state-of-the-art performance on four large benchmarks (DAVIS, YouTube-VOS, SegTrack-v2, and YouTube-Objects), demonstrating the effectiveness of the proposed approach.
\end{abstract}

\section{Introduction}
Video object segmentation (VOS) in semi-supervised setting aims to segment the class-agnostic objects or instances from the background according to the annotation in the first frame, which has been widely applied to video editing, automatic driving, etc.
Tremendous progress \cite{johnander2019generative,caelles2017one,wug2018fast,wang2019fast} has been made with deep learning methods in recent years,
most of which directly embed the whole frame image or propagate the previous mask into current frame.
However, it is still challenging due to the background noise, object missing, or severe occlusion in real world scenarios.

\begin{figure}[t]
% \vspace{-10pt}
\centering
\includegraphics[width=0.48\textwidth]{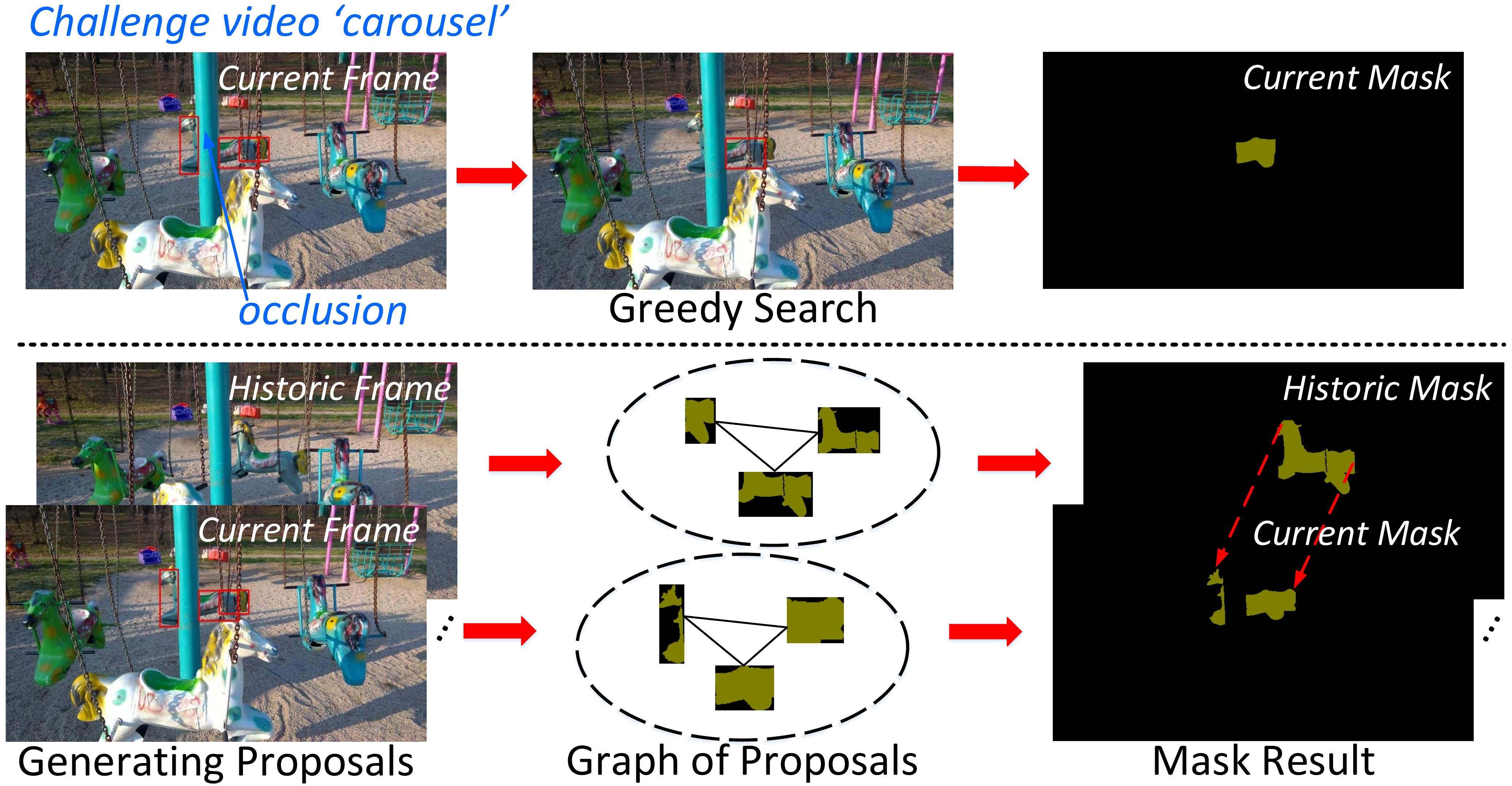}
\caption{Different from previous detection based methods that generally utilize a greedy strategy to choose the best proposal for segmentation, our spatiotemporal graph considers all proposals of each frame in the spatial domain and utilizes historic masks for refinement in the temporal domain, which provides better mask details.}
\label{fig:introduction}
\vspace{-10pt}
\end{figure}

To address such challenges, detection based schemes \cite{li2017video,luiten2018premvos} are proposed, which 
% restore missing objects or re-establish objects associations with bounding box (bbox) proposals.
restore missing objects or re-establish objects with bounding box proposals. These proposals of target objects are either generated individually in each frame by detectors like Mask R-CNN \cite{he2017mask}, or further merged with a few adjacent neighboring frames \cite{luiten2018premvos}.
Although they are effective in object missing and occlusion scenarios,
these approaches rely on a greedy search scheme that extracts the best proposal in a frame, as shown in the upper part of Figure \ref{fig:introduction}, resulting in a strong dependence not only on the proposal quality, but also on a reliable Re-ID network \cite{li2017video} for proposal selection. 
Since the local patch details may be contained in all proposals scattered across the frame, instead of choosing the best proposal by a greedy search scheme, we argue that one should leverage the rich contexts of all proposals to reconstruct a more accurate mask.

Recently, graph neural network (GNN) \cite{scarselli2008graph,gilmer2017neural,cheng2020spatio,cheng2020graph,liu2020jointly} is recognized as a promising approach in sequential information processing. It takes advantage of aggregating the information with node-wise correlation establishment.
% It is not the first time that GNN is used in VOS task. AGNN~\cite{wang2019zero} represents frame as node, and utilize relations between arbitrary frame pairs as edges to capture higher-order understanding of video content only on the temporal dimension. GO-VOS \cite{haller2019spacetime} is proposed to represents pixel as node which is linked by optical flow, which may loss instance-level relations. 
Previous GNN based methods \cite{wang2019zero,haller2019spacetime,bao2018cnn} in the VOS task either represent image frames as nodes to explore their temporal correlations, or represent pixels as nodes that may lose instance-level relations. They may also fail to recover the spatial mask details in local contexts.
To better capture the local patch details for more accurate mask reconstruction, different from them, we take masks of all detection based proposals in the same frame as nodes to construct a spatial graph, and correlate current frame result to the previous frame masks to build a temporal graph. Such a joint graph neural network is capable of not only aggregating the scattered instance-level mask details of current frame (in the spatial domain), but also capturing the temporal correlations with historic masks (in the temporal domain).

In this paper, to better capture the local patch details from spatial information of the current frame and capture motion clues with temporal information from the previous frames, we propose a novel spatiotemporal graph neural network (STG-Net) for video object segmentation. Specifically, we construct a fully-connected spatial graph on the mask proposals of current frame to establish intra-object proposals relationship, and propose a temporally-connected graph to link the historic masks, as shown in the bottom part of Figure \ref{fig:introduction}. In a spatial graph, we develop an edge weight strategy to represent the correlation between two instance-level mask proposals by considering their feature similarity. After spatial graph updating, we design a score function based on motion estimation and mask propagation to choose the best reconstructed mask from all nodes for each object in current frame. Then a temporal graph is developed to correlate the chosen mask with historic masks of previous frame for mask refinement, which can also be regarded as a reconstruction process. Therefore, the mask is reconstructed in both spatial and temporal domains to produce a more accurate segmentation that recovers detailed contexts of objects.
% In this paper, we propose a novel spatiotemporal graph neural network (STG-Net) for video object segmentation. Specifically, we construct a fully-connected spatial graph on the generated proposals of current frame to establish intra-object proposals relationship, and propose a temporally-connected graph to link the historic masks, as shown in the bottom part of Figure \ref{fig:introduction}. 
% In a spatial graph, we specially develop an edge weight strategy to represent the correlation between two instance-level mask proposals by considering their feature similarity.
% After spatial graph updating, we design a score function based on motion estimation and mask propagation to choose the best reconstructed mask from all nodes.
% Then a temporal graph is developed to link the chosen mask of current frame with historic masks for mask refinement, which can also be regarded as a reconstruction process. 
% We first retrieve the mask contexts from historic masks by utilizing a memory network. Then the chosen mask of current frame is refined with such previous information to obtain the final result. Therefore, the mask is reconstructed in both spatial and temporal domains to produce a more accurate segmentation that recovers detailed contexts of objects. 

The contributions of our work are summarized as follows:
\begin{itemize}
    \item We propose a novel VOS method named STG-Net based on a spatiotemporal graph to recover the local patch details in an instance level. With the cooperation of spatial and temporal graph networks, STG-Net has sufficient capacity to aggregate detailed mask contexts for more accurate mask reconstruction. To the best of our knowledge, it is the first time to take advantage of both spatial and temporal correlations with GNN in VOS task.
    %  for mask reconstruction
    \item 
    Instead of searching the best proposal in a greedy manner, our spatial graph network takes all object proposals into consideration with an edge weight strategy, which is measured by the feature similarity of a proposal pair. It helps to aggregate mask details from scattered locations. A score function is then employed to choose the reconstructed mask from spatial graph by considering both motion estimation and mask propagation.
    \item We develop a temporal graph based on the chosen masks from spatial graph along the time dimension. For each node, we utilize a memory network to retrieve mask contexts from the historic masks, and use them to refine the current mask with a temporal graph network.
\end{itemize}

Experimental results show that the proposed STG-Net achieves state-of-the-art performance on DAVIS, YouTube-VOS, SegTrack-v2, and YouTube-Objects datasets without online learning on the annotation of the first frame. The superior visual results show better mask details than others, which demonstrates the effectiveness of our method in handling the challenging occluded and missing objects.

\section{Related Works}
% In this section, we briefly summarize recent researches including semi-supervised video object segmentation, graph neural networks, and memory networks.
\noindent \textbf{Semi-Supervised Video Object Segmentation.}
Semi-supervised video object segmentation can be roughly classified into three categories: matching-based, propagation-based, and detection-based methods. Matching-based methods \cite{caelles2017one,voigtlaender2017online,zeng2019dmm} generally utilize the given mask in the first frame to extract appearance information for objects of interest, which is then used to find similar objects in succeeding frames. 
Some works \cite{caelles2017one,voigtlaender2017online} trained a parent network on still images and then fine-tuned the pre-trained work with one-shotonline learning. 
Embedding approaches \cite{chen2018blazingly,hu2018videomatch} mapped pixels to group the pixels of same object, and there are methods \cite{voigtlaender2019feelvos,wang2019ranet} extend from them for multiple object segmentation with correlation operation.
Propagation-based methods \cite{wug2018fast,johnander2019generative,xu2019mhp} utilize temporal information to refine masks propagated from preceding frames.
The above two methods mainly depend on the robustness of the feature extractor to segment the foreground object in the whole image where much background noise may be induced. Different from them, detection-based methods \cite{li2017video,wang2019fast,luiten2018premvos} first detect the best bounding box of each object in a frame, then crop out the target and input it into a segmentation model. Although it can decrease background noises and improve the performance of segmentation, it relies on the quality of the generated bounding box of each object. To avoid losing local details, our method aggregates the mask contexts of all proposals in the same frame to automatically reconstruct a more accurate mask.

\noindent \textbf{Graph Neural Networks.}
Graph neural network (GNN) \cite{scarselli2008graph} is an extension for recursive neural networks and random walk based models for graph structured data. \cite{gilmer2017neural,li2018deeper} further adapted GNN to sequential outputs with a learnable message passing module.
Generally, for each node $v$ in a graph, the updating process includes two steps: message aggregation and hidden state update. For the updating of $l$-th iteration, the node $v$ first aggregates messages from its neighbors $\mathcal{N}(v)$ into a single message $\bm{m}^v$ and then updates the hidden state $h^v$ itself with $\bm{m}^v$, it is according to:
\begin{equation}
    \bm{m}^v=F(h^u, u\in \mathcal{N}(v)), \quad (h^v){'}=U(h^v, \bm{m}^v),
\end{equation}
where $F(\cdot),U(\cdot)$ are the functions to update the message and hidden state. 
Different from \cite{haller2019spacetime,wang2019zero} that take a naive GNN to the VOS task and treat frames as nodes for temporal contexts exploring, our spatial graph is constructed with object proposals of each frame with a tailored edge weight strategy.

\begin{figure*}[t]
\centering
\includegraphics[width=1.0\textwidth]{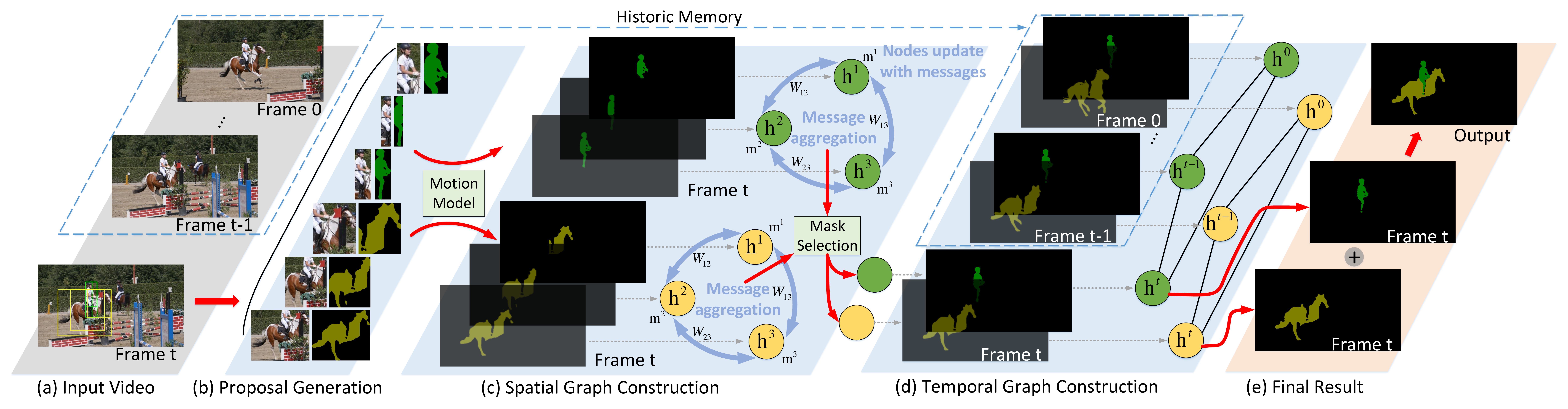}
\caption{The framework of our proposed scheme. Given a video frame $t$, we first obtain bbox proposals by an offline detection model and generate corresponding masks by an offline segmentation model. Then we classify each proposal with our motion model to construct intra-object spatial graphs, where each node is a mask proposal. During the spatial graph updating, each node aggregates mask contexts from its neighbors and updates itself to reconstruct more accurate mask. We further design a mask selection function to choose the best reconstructed mask from all nodes in graph. At last, we construct a temporal graph to retrieve historic masks of previous frames for refinement, and obtain the final result of frame $t$.}
\label{fig:pipeline}
\vspace{-10pt}
\end{figure*}

\noindent \textbf{Memory Networks.}
Memory networks \cite{vaswani2017attention,sukhbaatar2015end} have external memory where information can be further written and read. Given an input, the information is separately embedded into key and value feature maps, where the key feature maps are used to address relevant memories whose corresponding value feature maps are returned. 
% The retrieved values are then send to another network to obtain the output. 
% For structure, our temporal graph takes bounding box based image/mask as key/value for retrieval with only one network. Instead, STM takes whole image as input, and utilizes an encoder-decoder structure to embed image into two feature spaces named as key/value. For goal, we aim to propagate previous masks on edges for current mask refinement. And STM is to predict mask for each input.
Different from the embedding process \cite{oh2019video,lu2020video} for value features maps, we take the mask result of each frame as value, to retrieve historic masks for refining the current mask in temporal graph.

\section{Spatiotemporal Graph Neural Network}
In this section, we present our method STG-Net, as illustrated in Figure \ref{fig:pipeline}. Given a video frame, we first utilize a class-agnostic detection model to generate object bounding box (bbox) proposals and a class-agnostic segmentation model to segment corresponding masks, which is performed offline. Then, we design a motion model to classify the proposals, and take the proposals of the same object as nodes to construct a spatial graph. After spatial graph updating, we choose the best reconstructed mask from all nodes, and capture the temporal information by using a temporal graph that utilizes historic masks of previous frames for refinement. We pre-process the proposal generation offline, thus our whole framework can be end-to-end training.

\subsection{Proposal Generation}
We first generate possible object bbox proposals of a video frame by using an offline detection method Mask R-CNN \cite{he2017mask}, where each bbox proposal contains different local patch details of objects. As VOS task only considers foreground objects, we change the number of categories in Mask R-CNN into only one class to make the bbox proposals class-agnostic. Specifically, we extract object bboxs with pre-defined thresholds of detection confidence and non-maxinum suppression to keep the most possible bboxs. Given a video frame $\bm{I} \in \mathbb{R}^{3 \times H\times W}$, we denote the extracted bbox proposal as $b^v = (x_{\text{min}}^v, y_{\text{min}}^v, x_{\text{max}}^v, y_{\text{max}}^v)$, where $v$ is the $v$-th proposal of detection results in current frame.
To further segment its corresponding mask $\bm{M}^v$, we employ Deeplabv3+ \cite{chen2018encoder} network with ResNet101 \cite{he2016deep} backbone offline. 
Since objects tend to move smoothly through space in time, we use optical flow generated by FlowNet2.0 \cite{ilg2017flownet} as a direct source of information, to estimate a rough mask $\bm{Q}$ for current frame by a warp operation \cite{khoreva2017lucid}. And we take $\bm{Q}$ as an additional input besides the cropped RGB image $\bm{I}^v$ which is based on bbox $b^v$, to guide the segmentation module to produce more accurate mask.

\subsection{Spatial Graph Construction}
Different proposals may contain different local patch details, we construct spatial graphs in Figure \ref{fig:pipeline}(c) to aggregate the mask contexts of all proposals to enrich the segmentation information. To handle the class-agnostic mask proposals $\bm{M}^v$, we first introduce a motion model to prepare for proposal-wise classification, then our spatial graph can be built with intra-object proposals for contexts sharing. We construct our spatial graph of each object as follows.

\noindent \textbf{Preparation.}
% After getting the previous frame bboxs of object $o$ as history, we aim to find the most closely proposals in current frame as the intra-object ones to build the graph. 
Before constructing the spatial graph of object $o$, we first group the $o$-class proposals from all proposals in current frame by classification.
As object smoothly moves across the video, we can utilize its mask results in previous frames and generate corresponding bboxs $\{b\}$ and center points $\{c\}$ as motion history. Based on such history memory, we can predict a bbox $p$ as objective probability location in current frame based on the prior knowledge of previous frame bboxs, then take the closest proposal $b^v$ into object $o$ class.
Since each bbox mainly depends on the characters of size $s_{t}$ and center point $c_{t}$ where $s_{t}$ is composed of the height and weight of bbox, $t$ denotes the time-step, we develop a motion model to estimate the center point of $p$ based on the previous $n$ steps movements by:
\begin{equation}
    c_{t} = c_{t-1} + \frac{1}{n}\sum_{r=t-n}^{t-1}(c_{r}-c_{r-1}),
\end{equation}
where the second term means the average velocity. The bbox size $s_{t}$ of $p$ can also be estimated as $s_{t}=\frac{1}{n}\sum_{r=t-n}^{t-1}s_r$, since most object sizes change smoothly in video sequence. Therefore, the predicted objective probability bbox $p$ of object $o$ in current frame $t$ is composed of $(c_{t},s_{t})$.
% Then we need to search the proposals of object $o$ among $b^v$ with the probability objective bbox $p^o$ to filter out the inter-object noise.
Given a bbox proposal $b^v$, we first calculate the intersection over union (IoU) scores between the area of $b^v$ and the area of $p$ for all object. Then we rank the IoU scores to find the object class of the highest one to classify the bbox $b^v$.
If the top ranked score refers to the object $o$, we add the bbox $b^v$ into the object $o$-class.
After getting the proposal index $v,v=1,...,N$ of object $o$-class, we past corresponding mask into a void image to rebuild back a full mask $\bm{M}^v \in \mathbb{R}^{H\times W}$. 

\noindent \textbf{Graph construction.}
To recover the local patch details in intra-object proposals, we construct a fully-connected spatial graph of object $o$ with masks $\{\bm{M}^{v}\}$ as node, to propagate beneficial information for mask reconstruction. Each node aggregates the mask contexts from its neighbors to reconstruct the mask itself. Generally, the correlation between different nodes are not always the same, as the proposals which have closer position and similar representation tends to be more relevant. 
To selectively propagate the mask information more between the relevant nodes while reduce the noise between less relevant ones, we define an edge weight $\bm{W}_{vu}$ on the edge between node $v$ and $u$, which can be formalized as follows:
\begin{equation}
    \bm{W}_{vu} = \left\{  
             \begin{array}{cl}
              \alpha {\rm cos}(\bm{X}^v, \bm{X}^u)+\beta  \text{IoU}(b^{v}, b^{u}),  & v \neq u  \\  
              0,  & v = u  \\
             \end{array},
\right. 
\label{eq:alpha}
\end{equation}
% \begin{equation}
%     \mathcal{C}(b^{v}, b^{u}) = {\rm cos}(f_{\theta}(\bm{I}^v), f_{\theta}(\bm{I}^v))
% \end{equation}
where the first item is the cosine similarity to measure the correlation between features $\bm{X}^v$ and $\bm{X}^u$ extracted by a learnable CNN for the proposals $b^v$ and $b^u$, respectively. $\alpha$ and $\beta$ control the ratio between the feature similarity score and the IoU score, and we set $\bm{W}_{vv}$ to 0 to avoid self-enhancing. This weight strategy also helps to reduce the influence on the edge between the wrong classified proposal from motion model and the correct one as their similarity score will be much smaller.

\noindent \textbf{Graph updating.}
% For the updating of our intra-object spatial graph, masks of nodes are propagated on the edges to its neighbors for message aggregation, and then each node updates itself based on the neighborhood messages to reconstruct more contextual segmentation result than before.
For node $v$ in graph, the updating process contains two steps: 1). Mask message aggregation: To attach the mask information from other nodes $u$, node $v$ first aggregates the mask message $\bm{m}^v$ from its neighbors by a weighted summation with the edge weight $\bm{W}_{vu}$; 2). Node mask update: Then node $v$ updates the state $h^v$ itself with the aggregated messages $\bm{m}^v$ to reconstruct more contextual segmentation result.
In details, we first initial the state $h^v$ of node $v$ with the mask proposal $\bm{M}^{v}$, and defined its neighbor sets as $\mathcal{N}(v)$. During the graph updating, it first aggregates messages $\bm{m}^v$ from neighbors $\mathcal{N}(v)$ by function $F(\cdot)$ as follow:
\begin{equation}
    \bm{m}^v = F(h^u) = \sum_u \bm{W}_{vu}h^u, u\in \mathcal{N}(v).
\end{equation}
Then node $v$ reconstructs the mask of its state with the aggregated mask message $\bm{m}^v \in \mathbb{R}^{H\times W}$ by:
\begin{equation}
    (h^v)^{'} = U(h^{v}, \bm{m}^v) = \frac{(1-\bm{W}_{vv})h^{v}+\bm{m}^v}{1+\sum_u \bm{W}_{vu}} , u\in \mathcal{N}(v),
\end{equation}
where $U(\cdot)$ is the function to update the mask state, and $(1+\sum_{u}\bm{W}_{vu})$ is used to normalize the mask result.
Specially, we iterative the graph updating process for several steps.
To avoid over-smoothing problem \cite{li2018deeper} in graph nodes, we only conduct less than three iterations and keep the edge weight $\bm{W}_{vu}$ unchanged during the graph iterative updating. 
% Graph updating details are shown in Algorithm \ref{alg:algorithm}.
At last, for node $v$, we get the binary reconstructed mask $\widehat{\bm{M}}^{v} = (h^v>thr)$ by a threshold $thr$, and it recovers the local patch details from intra-object proposals.

\subsection{Temporal Graph Construction}
After getting masks $\{\widehat{\bm{M}}^{v}\}$ in spatial graph, we design a score function to choose the best mask of object $o$ from all nodes. Then, we refine it using the temporal information of the same instance by a temporal graph in Figure \ref{fig:pipeline}(d).

\noindent \textbf{Mask selection.}
We define the score function as follows:
\begin{equation}
    \mathcal{S}(\widehat{\bm{M}}^{v}|(p,\bm{Q})) = \lambda_1 \text{IoU}(\mathcal{B}(\widehat{\bm{M}}^{v}), p) + \lambda_2 \frac{\widehat{\bm{M}}^{v} \cap \bm{Q}}{\widehat{\bm{M}}^{v} \cup \bm{Q}},
\label{eq:lambda}
\end{equation}
where $\mathcal{B}(\cdot)$ is the function to extract the bbox of the reconstructed mask $\widehat{\bm{M}}^{v}$, and $\bm{Q}$ is the warped mask from previous frame with optical flow. Our mask selection score contains two parts: 1) bbox IoU score between the bbox of current segmented object and predicted probability bbox $p$ from motion model, which stands for the measurement of object motion estimation; 2) the intersection area over union area score between the current estimated mask and the warped mask, which represents the performance of mask propagation. $\lambda_1$ and $\lambda_2$ are the parameters to control the ratio of these two scores. We choose the best mask index $v$ with $\max_{v} \mathcal{S}(\widehat{\bm{M}}^{v}|(p,\bm{Q}))$, and denote the chosen mask as current frame segmentation result $\widehat{\bm{M}} \in \mathbb{R}^{H\times W}$ of object $o$.

\noindent \textbf{Graph construction.}
% Given the chosen mask $\widehat{\bm{M}}$, we further refine it using the temporal information of the same instance. 
Inspired by differential memory networks \cite{vaswani2017attention,sukhbaatar2015end}, we utilize the chosen mask of object $o$ in current frame and the previous frame mask results as nodes, to construct a temporal graph. This graph is designed to refine the mask of the current frame by aggregating historic masks as memories and is evolved by adding new nodes over time. At frame $t$, there are $t+1$ nodes with mask $\widehat{\bm{M}}^r,r\leq t$ including the first frame $0$. 
We first crop the image $\bm{I}$ based on $\mathcal{B}(\widehat{\bm{M}}^r)$, and resize it to $\bm{I}^r \in \mathbb{R}^{3 \times H_1 \times W_1}$. Then, we
extract its embedding features by a learnable CNN as the key feature maps $\bm{K}^r \in \mathbb{R}^{C \times H_1 \times W_1}$. We take the cropped $\widehat{\bm{M}}^r$ as the corresponding value map. 
% We extract the memory key maps of each step $\bm{K}^r=k_{\theta}(\bm{I}^r) \in \mathbb{R}^{H \times W \times C}$ by a CNN $k_{\theta}$ where $\bm{I}^r$ is the cropped image based on the bbox $\mathcal{B}(\widehat{\bm{M}}^r)$,  and take mask $\widehat{\bm{M}}^r$ as value map $\bm{V}_r$. 
The similarity between key feature maps of the current and previous frames are computed to determine when-and-where to retrieve relevant previous value maps from. Therefore, every pixel of each previous frame value map can be utilized to construct a new self-predicted mask for current frame based on such similarities. These constructed masks are taken as neighborhood messages $\bm{m}^t$ for current frame node $t$ to update its mask $h^t$. 

The main differences of our memory network compared to STM \cite{oh2019video} are:
For structure, our temporal graph takes bbox based image/mask as key/value with only one network. Instead, STM takes whole image as input, and utilizes an encoder-decoder structure to embed image into two feature spaces named as key/value. For goal, we aim to propagate previous masks on edges for current mask refinement. And STM is to predict mask for each input.

\begin{figure}[t]
\centering
\includegraphics[width=0.47\textwidth]{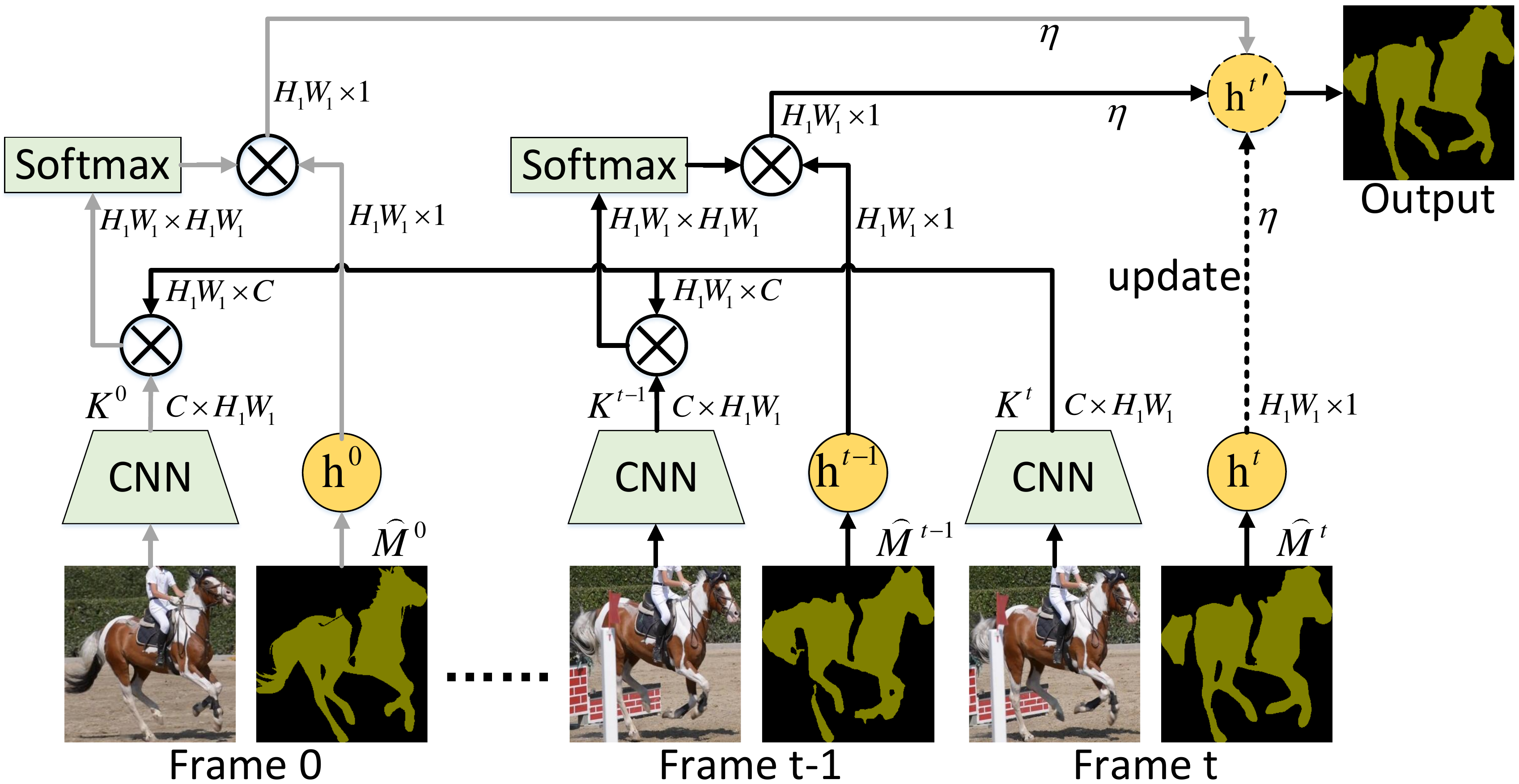}
\caption{Illustration of temporal graph updating process at frame $t$, where `$\bigotimes$' denotes the matrix multiplication.}
\label{fig:temporal}
\vspace{-10pt}
\end{figure}

\noindent \textbf{Graph updating.}
In details, node $t$ first initials its state $h^t$ with mask $\widehat{\bm{M}}^t$, and then retrieves the mask memories from previous $t$ frames including the first frame $0$ as:
\begin{equation}
    (\bm{m}^t)_i = F(h^r) = \sum_{r=0}^{t-1} \frac{\sum_j{\text{exp}((\bm{K}^t)_i \odot (\bm{K}^r)_j) (h^r)_j}}{\sum_j{\text{exp}((\bm{K}^t)_i \odot (\bm{K}^r)_j)}}, 
\end{equation}
where $i,j$ are the location index and $(\bm{K}^t)_i \in \mathbb{R}^{C\times 1}$, $\odot$ denotes the dot-product. Therefore, the refinement process of frame $t$ by using the mask memories can be seen as a process of reconstruction, which is formulated by:
\begin{equation}
    (h^t)^{'}  = U(h^t, \bm{m}^t) = \eta(h^t + \bm{m}^t),
\end{equation}
where $\eta=1/(t+1)$ is used to normalize the mask result. Details of temporal graph updating at frame $t$ are shown in Figure \ref{fig:temporal}. Note that our temporal graph only updates within one iteration, and the refined output mask $\widehat{\bm{M}} \in \mathbb{R}^{H \times W}$ of current frame is also obtained by $\widehat{\bm{M}} = (h^t>thr)$.

\subsection{Network Structure and Loss Function}
For the feature extractor in spatial graph, we use a ResNet101 backbone of which the weights are initialized from the released model of RVOS \cite{ventura2019rvos}, and obtain the proposal features from its last layer. As for memory key maps, we use the stage-4 feature map of a ResNet50 which is finetuned online. During the training, as for the reconstructed mask $\widehat{\bm{M}}^{v}$ in spatial graph, we utilize the ground truth pair $(\rm{b}, \textbf{M})$ to choose the best mask instead of $(p,\bm{Q})$. To make our model end-to-end trainable with the mask selection operation, we develop an additional distance loss to backpropagate the gradient for all reconstructed masks. Our total loss function $\mathcal{L}$ is formulated as:
% \begin{align}
%     \mathcal{L}(M_t^o) = & - g_t^o {\rm log}(\sigma(M_t^o)) - (1-g_t^o){\rm log}(1-\sigma(M_t^o)) \nonumber\\
%     + & \gamma\sum_{i} ||{ S}(r_t^{i_o}|(b_{gt}, M_{gt}))-{S}(r_t^{i_o}|(b_{gt}, M_{gt}))||
% \end{align}
\begin{align}
    \mathcal{L}(\widehat{\bm{M}}) = & \gamma\sum_{v} |{\mathcal{S}}(\widehat{\bm{M}}^{v}|({\rm{b}, \textbf{M}}))-{\mathcal{S}}(\widehat{\bm{M}}^{v}|({p, \bm{Q}}))| \nonumber\\
    - & {\rm \textbf{M}}{\rm log}(\sigma(\widehat{\bm{M}})) - (1-{\rm \textbf{M}}){\rm log}(1-\sigma(\widehat{\bm{M}})),
\label{eq:loss}
\end{align}
where $\sigma(\cdot)$ denotes the sigmoid function, and $\gamma$ is the hyperparameter to balance the two losses. 
% During the training, the similarity extractor will generate more accurate score result.
% During the test phase, our model just directly choose the best reconstructed mask based on the warped mask.

\section{Experiments}
% In this section, we compare our method with a wide range of recent competitors on DAVIS2017 \cite{pont20172017}, YouTube-VOS \cite{xu2018youtube}, SegTrack-V2 \cite{li2013video} and YouTube-Objects \cite{prest2012learning} datasets. We first describe the implementation details and
% the datasets used for video object segmentation before we present evaluation metrics, quantitative and qualitative results, and the analysis of our proposed network.
\subsection{Datasets}
\textbf{DAVIS.} DAVIS2016 and DAVIS2017 \cite{pont20172017} are widely used to evaluate VOS methods, where the former one focuses on object-level segmentation and the latter one is challenging in multiple objects which correspond to different targets. 
It consists of 60 videos in training set and 30 videos in the validation set. It also provides extra test-dev data with 30 challenging videos, which contains some similar objects in the same videos and object occlusion or missing in the continues frames. 
And there are three evaluation metrics: region similarity $\mathcal{J}$, the intersection over union of the estimated segmentation and the ground truth mask; contour accuracy $\mathcal{F}$: F-measure between the contour-based precision and recall; and global mean value $\mathcal{G}$: average score of $\mathcal{J}$ and $\mathcal{F}$.

\noindent \textbf{YouTube-VOS.} YouTube-VOS \cite{xu2018youtube} is the latest large-scale dataset which consists of 4453 videos annotated with multiple objects. The validation set contains 474 videos including 91 object categories, and it has separate measures for 65 of seen and 26 of unseen object categories. Like DAVIS dataset, we adopt $\mathcal{J}$, $\mathcal{F}$ and $\mathcal{G}$ for evaluation.

\noindent \textbf{Others.} The SegTrack-v2 dataset \cite{li2013video} consists of 14 test video sequences with 24 objects. YouTube-Objects \cite{prest2012learning} comprises 126 video sequences which belong to 10 object categories. Following the protocol, we use metric $\mathcal{J}$ to measure the segmentation performance.

\begin{table*}
\centering
\begin{tabular}{c|c|cccccccccccc}
\hline
\multirow{2}*{Method} & \multirow{2}*{OL} & \ & \multicolumn{3}{c}{DAVIS2017 test-dev} & \ &  \multicolumn{3}{c}{DAVIS2017 val} & \ & \multicolumn{3}{c}{DAVIS2016 val}\\ \cline{4-6} \cline{8-10} \cline{12-14}
~ & ~ & \ & $\mathcal{J}_\mathcal{M}$ & $\mathcal{F}_\mathcal{M}$ & $\mathcal{G}_\mathcal{M}$ & \ & $\mathcal{J}_\mathcal{M}$ & $\mathcal{F}_\mathcal{M}$ & $\mathcal{G}_\mathcal{M}$ & \ & $\mathcal{J}_\mathcal{M}$ & $\mathcal{F}_\mathcal{M}$ & $\mathcal{G}_\mathcal{M}$  \\ \hline
OSMN \cite{yang2018efficient} & \xmark & \ & 37.7 & 44.9 & 41.3 & \ & 52.5 & 57.1 & 54.8 & \ & 74.0 & 72.9 & 73.5 \\
SiamMask \cite{wang2019fast} & \xmark & \ & 40.6 & 45.8 & 43.2 & \ & 54.3 & 58.5 & 56.4 & \ & 71.7 & 67.8 & 69.8 \\
FAVOS \cite{cheng2018fast} & \xmark & \ & 42.9 & 44.2 & 43.6 & \ & 54.6 & 61.8 & 58.2 & \ & 82.4 & 79.5 & 81.0 \\
RVOS \cite{ventura2019rvos} & \xmark & \ & 47.9 & 52.6 & 50.3 & \ & 57.5 & 63.6 & 60.6 & \ & - & - & -  \\
% OSVOS \cite{caelles2017one} & \cmark & \ & 47.0 & 54.8 & 50.9 & \ & 56.6 & 63.9 & 60.3 & \ & 79.8 & 80.6 & 79.2 \\
AGAME \cite{johnander2019generative} & \xmark & \ & 49.2 & 55.3 & 52.3 & \ & 68.5 & 73.6 & 71.0 & \ & 81.5 & 82.2 & 81.9 \\
RGMP \cite{wug2018fast} & \xmark & \ & 51.3 & 54.4 & 52.8 & \ & 64.8 & 68.6 & 66.7 & \ & 81.5 & 82.0 & 81.8 \\
AGSS \cite{lin2019agss} & \xmark & \ & 51.5 & 57.1 & 54.3 & \ & 63.4 & 69.8 & 66.6 & \ & - & - & - \\
FEEL \cite{voigtlaender2019feelvos} & \xmark & \ & 51.2 & 57.5 & 54.4 & \ & 65.9 & 72.3 & 69.1 & \ & 80.3 & 83.1 & 81.7 \\
RANet \cite{wang2019ranet} & \xmark & \ & 53.4 & 57.2 & 55.3 & \ & 63.2 & 68.2 & 65.7 & \ & \textbf{85.5} & 85.4 & 85.5 \\
AGSS$^*$ \cite{lin2019agss} & \xmark & \ & 54.8 & 59.7 & 57.2 & \ & 64.9 & 69.9 & 67.4 & \ & - & - & - \\
FEEL$^*$ \cite{voigtlaender2019feelvos} & \xmark & \ & 55.1 & 60.4 & 57.8 & \ & 69.1 & 74.0 & 71.5 & \ & 81.1 & 82.2 & 81.7 \\ \hline
\textbf{Ours} & \xmark & \ & \textbf{59.7} & \textbf{66.5} & \textbf{63.1} & \ & \textbf{71.5} & \textbf{77.9} & \textbf{74.7} & \ & 85.4 & \textbf{86.0} & \textbf{85.7} 
\\ \hline
\end{tabular}
\caption{Quantitative comparison of state-of-the-art methods on DAVIS2016 validation, DAVIS2017 validation and test-dev sets. $\mathcal{M}$ denotes the mean value. ``OL" indicates online learning with the annotation of the first frame. $^*$ indicates the use of YouTube-VOS for pre-training.}
\label{tab:davis}
\vspace{-8pt}
\end{table*}

\subsection{Implementation Details}
To adapt the Mask R-CNN network to generate class-agnostic foreground object bboxs offline, we first train it on COCO dataset with the pre-trained weights on ImageNet. Then we finetune it on DAVIS2017 and YouTube-VOS respectively. In testing phase, we set detection confidence as 0.05 and non-maxminum suppression as 0.6.
To feed the bbox proposals to segmentation module Deeplabv3+, we crop the bbox of the four channel input by using the spatial information of the annotation with a margin ratio 0.15. Then we resize the cropped data into $512 \times 512$, jitter the image color. Similar to the training process of Mask R-CNN, we first pre-train Deeplabv3+ on COCO dataset, and then train it on DAVIS and YouTube-VOS with learning rate 1e-5 for 100 epochs respectively.

To train our spatiotemporal graph together with the two feature extractors, we set Adam optimizer with learning rate 0.1 which reduces by power of 0.9 for every 10 epochs. We adopt $n=10$ to store history in our motion mechanism. The balanced ratio $\alpha$ and $\beta$ in Eq.(\ref{eq:alpha}) are set to 0.7 and 0.3 for DAVIS and YouTube-VOS, 0.1 and 0.9 for SegTrack-v2 and YouTube-Objects. And we set the parameters $\lambda_1$ and $\lambda_2$ in Eq.(\ref{eq:lambda}) to 0.4 and 0.6 respectively for their relative importance. The $\gamma$ in Eq.(\ref{eq:loss}) is set to 100. The $thr$ is set to 0.2.All experiments are implemented on a single NVIDIA 1080Ti GPU. Our codes and trained model will be available online.

\begin{table}[t!]
\small
\centering
\begin{tabular}{c|c|cccc|c|c}
\hline
Method & OL & $\mathcal{J}_\mathcal{S}$ & $\mathcal{J}_\mathcal{U}$ & $\mathcal{F}_\mathcal{S}$ & $\mathcal{F}_\mathcal{U}$ & $\mathcal{G}_\mathcal{M}$ & FPS \\ \hline
OSMN & \xmark & 60.0 & 40.6 & 60.1 & 44.0 & 51.2 & 8.0 \\
DMM & \xmark & 58.3 & 41.6 & 60.7 & 46.3 & 51.7 & 12 \\
SiamMask & \xmark & 60.2 & 45.1 & 58.2 & 47.7 & 52.8 & \textbf{55} \\
RGMP & \xmark & 59.5 & - & 45.2 & - & 53.8 & 7 \\
OnAVOS  & \cmark & 60.1 & 46.6 & 62.7 & 51.4 & 55.2 & 0.1 \\
RVOS & \xmark & 63.6 & 45.5 & 67.2 & 51.0 & 56.8 & 24 \\
S2S & \xmark & 66.7 & 48.2 & 65.5 & 50.3 & 57.7 & 6 \\
DMM & \cmark & 60.3 & 50.6 & 63.5 & 57.4 & 58.0 & - \\
OSVOS & \cmark & 59.8 & 54.2 & 60.5 & 60.7 & 58.8 & 0.1 \\
S2S & \cmark & 71.0 & 55.5 & 70.0 & 61.2 & 64.4 & 0.1 \\
AGAME & \xmark & 66.9 & 61.2 & - & - & 66.0 & - \\
AGSS & \xmark & 71.3 & 65.5 & \textbf{75.2} & 73.1 & 71.3 & 12 \\ \hline
\textbf{Ours} & \xmark & \textbf{72.7} & \textbf{69.1} & \textbf{75.2} & \textbf{74.9} & \textbf{73.0} & 6\\ \hline
\end{tabular}
\caption{Quantitative comparison of state-of-the-art methods on YouTube-VOS validation set. `$\mathcal{S}$' and `$\mathcal{U}$' denote the seen and unseen categories. Specially, DMM \cite{zeng2019dmm}, OnAVOS \cite{voigtlaender2017online}, S2S \cite{xu2018youtube} and OSVOS \cite{caelles2017one} contain online learning.}
\label{tab:youtubevos}
\vspace{-10pt}
\end{table}

\subsection{Experimental Results}

\textbf{DAVIS.}
For experiments on DAVIS, we only train our STG-Net on DAVIS 2017 training set. We compare with a wide range of recent competitors both on the DAVIS2016 and DAVIS2017 datasets. Compared with approaches without online learning in Table \ref{tab:davis}, our method achieves the state-of-the-art performance and outperforms others over most evaluation criteria. Especially on DAVIS2017 test-dev set that contains much occlusion or object missing scenarios, our method achieves global mean value $\mathcal{G}_\mathcal{M}$ of 63.1, which gains a great margin of improvement than others. Compared to FEEL, we outperform them by 4.6\%, 6.1\% and 5.3\% on three metrics, respectively. Since DAVIS2016 only contains single object without challenge scenario, we perform the similar result to others but still gain improvement of 0.2.

\noindent \textbf{YouTube-VOS.}
For experiments on YouTube-VOS, we train our model on YouTube-VOS, and show the comparison with previous start-of-the-art approaches in Table \ref{tab:youtubevos}. Our method achieves a new state-of-the-art of global mean value $\mathcal{G}_\mathcal{M}$ 73.0 in terms of overall scores. Compared to the SOTA model AGSS, we outperform it by 1.7\%. Beside, we obtain a good trade off between the performance and running time. Compared with approaches without using online learning (like DMM, AGSS), our method achieves better performance on evaluation metrics. Compared with the approaches using online learning (like OSVOS, S2S), our method achieves faster FPS than these model, demonstrating that our method is more efficient.

\begin{table}
\scriptsize
\centering
\begin{tabular}{c|c|c||c|c|c}
\hline
Method & OL & SegTrack-v2 & Method & OL & YouTube-Objects \\ \hline
OSVOS & \cmark & 65.4 & OSMN & \xmark & 69.0 \\
% AGNN & \xmark & - & 70.8 \\
RGMP & \xmark & 71.1 & FEEL & \xmark & 78.9\\
DMM & \xmark & 76.7 & OnAVOS & \cmark & 80.5 \\ \hline
\textbf{Ours} & \xmark & \textbf{79.5} & \textbf{Ours} & \xmark & \textbf{84.1} \\ \hline
\end{tabular}
\caption{Quantitative comparison of state-of-the-art methods on SegTrack-v2 and YouTube-Objects datasets.}
\label{tab:other}
\vspace{-10pt}
\end{table}

\noindent \textbf{Others.} We test our STG-Net (trained on DAVIS2017) on SegTrack-V2 and YouTube-Objects datasets directly. Our method also achieves the state-of-the-art performance without online learning as shown in Table \ref{tab:other}.

\begin{figure*}[t]
\centering
\includegraphics[width=1.0\textwidth]{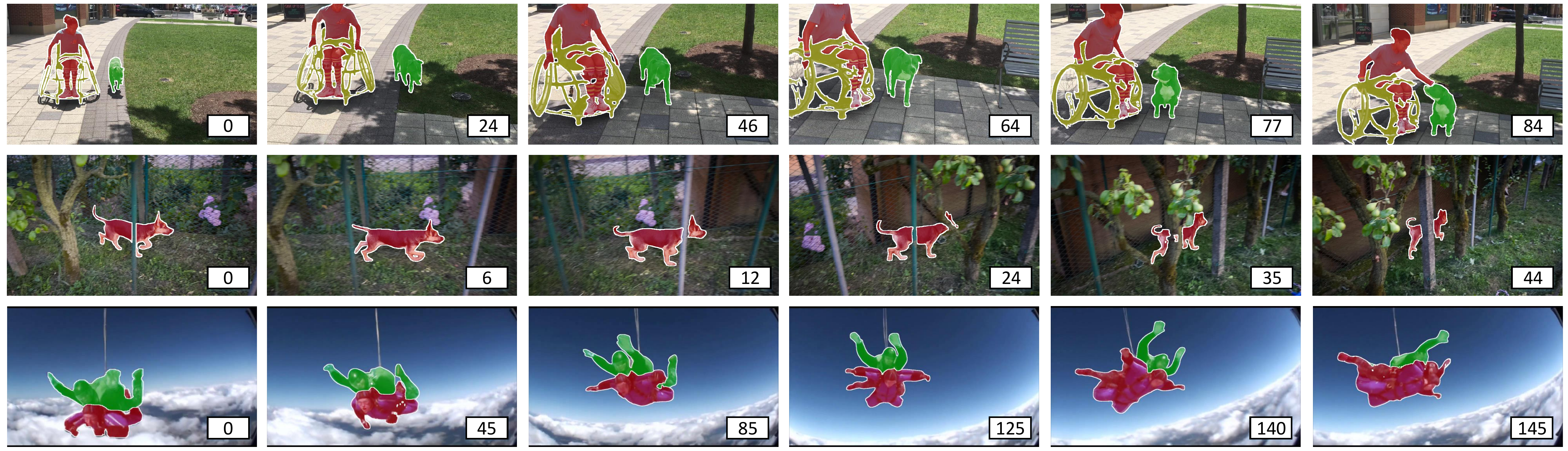}
\caption{Qualitative results of our method, where frames are sampled at the important moments (\textit{e.g.} multi-view or occlusion) for each video. From top to bottom, the sequences are ``girl-dog" in DAVIS2017, ``libby" in DAVIS2016 and ``0daaddc9da" in YouTube-VOS, respectively.}
\label{fig:sample}
\vspace{-10pt}
\end{figure*}

\begin{figure}[t]
\centering
\includegraphics[width=0.45\textwidth]{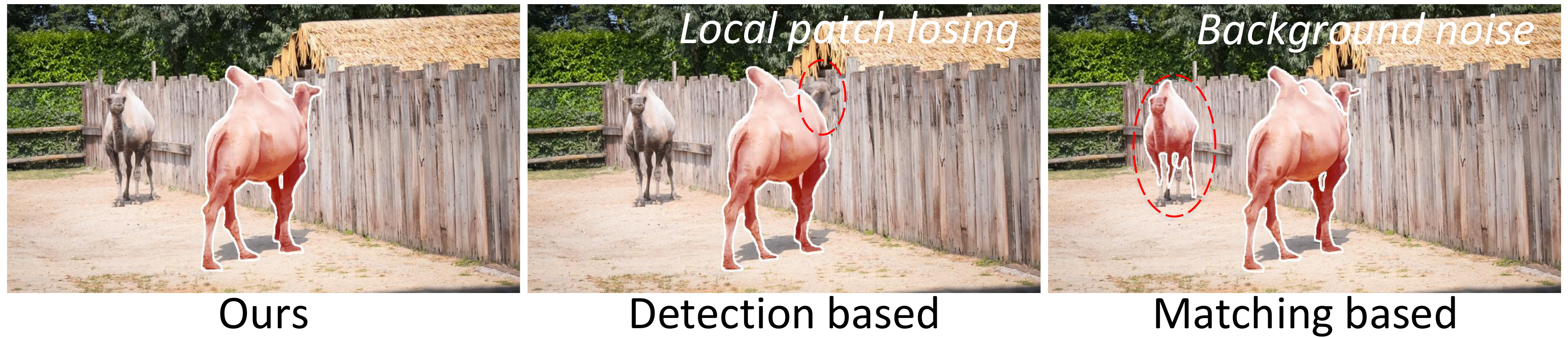}
\caption{Visual comparison with detection based \cite{luiten2018premvos} and matching based \cite{voigtlaender2019feelvos} methods.}
\label{fig:compare}
\vspace{-12pt}
\end{figure}

\subsection{Qualitative Visualization and Analysis}
Figure \ref{fig:sample} shows qualitative examples of our results, where we choose challenging videos from DAVIS2017, DAVIS2016, and YouTube-VOS datasets. Our method is robust to occlusions and complex motions. We also show the visual comparison in Figure \ref{fig:compare} where other detection based method loses the local patch details and matching based method wrongly segment the background visual similar object. Compared to them, our STG-Net reconstructs the mask result and provides better details. The reason is that our model recovers the local patch details in spatial domain by aggregating the contexts from all intra-object proposals. The temporal graph also helps to refine the mask by retrieving the historic mask memory. More qualitative results can be found in our supplymentary.

\subsection{Ablation Study}
We conduct thorough ablation study to analyze the effectiveness of different components and hyperparameters of STG-Net on DAVIS2017 test-dev set. Detailed results are shown in Table \ref{tab:ablation}. We start by a baseline model, which directly choose the best mask from the proposals with $\lambda_1,\lambda_2$ and achieves global mean value 53.2. 

\noindent \textbf{Effectiveness of the motion model.}
The baseline model computes the selection score in Eq. (6) using the previous bbox instead of the predicted $p$.
Considering the motion history in the previous frames, our motion model predicts a coarse position $p$ in current frame which is more accurate than the previous bbox. From the table, we find that it has the improvement of performance with 0.9.

\noindent \textbf{Effectiveness of the spatial graph.}
Our spatial graph reconstructs the mask by utilizing the local patch details from all proposals in current frame. It helps to deal with the occlusion and object missing problems. As shown in the table, our spatial graph construction makes the maximum improvement of 4.3 which recovers the local patch details among intra-object proposals. And the defined hyperparameters of $\alpha$, which controls the operation of the edge weighting, has another improvement of 0.8.

\noindent \textbf{Effectiveness of the temporal graph.}
Our temporal graph helps to refine the current frame mask result by using the previous frame masks, which can provides the temporal contextual information for better boundary results. As shown in table,
the temporal graph takes mask memory for refinement and has the second maximum improvement of 2.1.  The hyperparameters $\lambda_1$ in Eq. (6), which controls the operation of mask choosing, has another improvement of 0.6.

\noindent \textbf{How to choose the number of graph layer.}
We further investigate the performances on different iteration number $l$ of graph updating process. Although more graph layer will bring better performance, too much layers will result in oversmoothing problem. We find that the graph with 2-steps updating achieves the best result of 63.1.

\noindent \textbf{Comparison on different training strategy.}
We also do the experiments on different training strategy to investigate the benefits. Results show that online leaning with the annotations of the first frame brings the improvement of 1.2. Fine-tuning on the augmented train-set of DAVIS2017 makes the improvement of 2.2. And we add YouTube-VOS dataset for pre-training, it has another improvement of 2.9.

\begin{table}[t!]
\small
\centering
\begin{tabular}{cccccccccc}
\hline
\multicolumn{6}{c}{Settings} & \multirow{2}*{OL} & \multirow{2}*{FT} & \multirow{2}*{+ytb} & \multirow{2}*{$\mathcal{G}_\mathcal{M}$} \\ \cline{1-6}
Mo & SG & $\alpha$ & $l$ & $\lambda_1$ & TG & ~ & ~ & ~  & ~\\ \hline
\xmark & \xmark & -   & - & 0.5 & \xmark & \xmark & \xmark & \xmark & 53.2 \\
\cmark & \xmark & -   & - & 0.5 & \xmark & \xmark & \xmark & \xmark & 54.1 \\
\cmark & \cmark & 0.5 & 1 & 0.5 & \xmark & \xmark & \xmark & \xmark & 58.4 \\
\cmark & \cmark & 0.7 & 1 & 0.5 & \xmark & \xmark & \xmark & \xmark & 59.2 \\
\cmark & \cmark & 0.7 & 1 & 0.4 & \xmark & \xmark & \xmark & \xmark & 59.8 \\
\cmark & \cmark & 0.7 & 1 & 0.4 & \cmark & \xmark & \xmark & \xmark & 61.9 \\
\cmark & \cmark & 0.7 & 2 & 0.4 & \cmark & \xmark & \xmark & \xmark & \underline{63.1} \\
\cmark & \cmark & 0.7 & 3 & 0.4 & \cmark & \xmark & \xmark & \xmark & 62.3 \\ \hline
\cmark & \cmark & 0.7 & 2 & 0.4 & \cmark & \cmark & \xmark & \xmark & 64.3 \\
\cmark & \cmark & 0.7 & 2 & 0.4 & \cmark & \cmark & \cmark & \xmark & 66.5 \\
\cmark & \cmark & 0.7 & 2 & 0.4 & \cmark & \cmark & \cmark & \cmark & \textbf{69.4} \\ \hline
\end{tabular}
\caption{Ablation study evaluated on the DAVIS2017 test-dev set. `Mo' means the motion model,`SG' and `TG' mean the spatial and temporal graph. $\alpha$ controls edge weights where $\alpha=1-\beta$, and $\lambda_1$ controls the mask selection score where $\lambda_1=1-\lambda_2$. $l$ represents the spatial graph iteration number. `OL' indicates online learning and `FT' indicates fine-tuning with lucid \cite{khoreva2017lucid} augmentation. `+ytb' denotes pre-training on YouTube-VOS dataset.}
\label{tab:ablation}
\vspace{-10pt}
\end{table}

\section{Conclusion}
In this paper, we propose a spatiotemporal graph neural network (STG-Net) for video object segmentation.
By the cooperation of spatial and temporal graph networks, STG-Net has sufficient capacity to reconstruct more detailed masks.
Contrasted to the previous detection based approaches utilizing greed search strategy only in the current frame, the abundant use of both local patch details in spatial graph and time dimension relationships in temporal graph make STG-Net able to obtain a superior representation for the class-agnostic objects segmentation.
% Instead of utilizing
% Different from other object proposal based schemes, our graph takes all proposals of each frame into consideration instead of using a complex greedy search strategy. In spatial graph, we design an edge weight strategy to represent the proposal-wise correlation for messages aggregating. 
% And in temporal graph, we link the best reconstructed mask in time step for current mask refinement by using previous mask knowledge. 
Extensive experiments demonstrate that our method is robust to challenging scenarios thanks to our spatiotemporal graph, and outperforms state-of-the-art methods on all four benchmarks, even compared to online learning methods.

\clearpage
\noindent \textbf{Acknowledgements.} This work was supported in part by the National Natural Science Foundation of China under Grant (No.61972448, No.61602197), and in part by Domain Foundation of Equipment Advance Research of 13th Five-year Plan under Grant No.41412050801.

\bibliography{reference.bib}

\end{document}